%% file: top.tex
\PassOptionsToPackage{hyphens}{url}
\documentclass[conference, 10pt]{IEEEtran}
\IEEEoverridecommandlockouts
\usepackage{amsmath,amssymb,amsfonts}
\usepackage{algorithmic}
\usepackage{graphicx}
\usepackage{textcomp}
\usepackage{xcolor}
\usepackage{multirow}
\usepackage{algorithm}
\usepackage{hyperref}
\usepackage{balance}
\usepackage{booktabs}
\def\BibTeX{{\rm B\kern-.05em{\sc i\kern-.025em b}\kern-.08em
    T\kern-.1667em\lower.7ex\hbox{E}\kern-.125emX}}

\usepackage{eso-pic}
\newcommand{\InvitedBanner}{%
  \AddToShipoutPictureFG*{%
    \AtTextUpperLeft{%
}}}

\begin{document}

\title{Revolution or Hype? Seeking the Limits of Large Models in Hardware Design}

\author{
\IEEEauthorblockN{\large{Qiang Xu}}
\IEEEauthorblockA{\textit{The Chinese University of Hong Kong}\\
qxu@cse.cuhk.edu.hk\\[0.8em]
\large{Xi Wang}\\
\normalsize
\textit{Southeast University}\\
xi.wang@seu.edu.cn}
\and
\IEEEauthorblockN{\large{Leon Stok}}
\IEEEauthorblockA{\textit{IBM}\\
leonstok@us.ibm.com\\[0.8em]
\large{Grace Li Zhang}\\
\normalsize
\textit{TU Darmstadt}\\
grace.zhang@tu-darmstadt.de}
\and
\IEEEauthorblockN{\large{Rolf Drechsler}}
\IEEEauthorblockA{\textit{University of Bremen/DFKI}\\
drechsler@uni-bremen.de\\[0.8em]
\large{Igor L. Markov}\\
\normalsize
\textit{Synopsys}\\
imarkov@synopsys.com}
}

\maketitle
\InvitedBanner

\begin{abstract}
Recent breakthroughs in Large Language Models (LLMs) and Large Circuit Models (LCMs) have sparked excitement across the electronic design automation (EDA) community, promising a revolution in circuit design and optimization. Yet, this excitement is met with significant skepticism: Are these AI models a genuine revolution in circuit design, or a temporary wave of inflated expectations? This paper serves as a foundational text for the corresponding ICCAD 2025 panel, bringing together perspectives from leading experts in academia and industry. It critically examines the practical capabilities, fundamental limitations, and future prospects of large AI models in hardware design. The paper synthesizes the core arguments surrounding reliability, scalability, and interpretability, framing the debate on whether these models can meaningfully outperform or complement traditional EDA methods. The result is an authoritative overview offering fresh insights into one of today's most contentious and impactful technology trends.
\end{abstract}

\input{sections/01_Introduction}

\input{sections/02_Opportunities}

\input{sections/03_Challenges}

\input{sections/04_ExpertOpinions}
\input{sections/05_Future}
\input{sections/06_Conclusion}

\input{sections/Biographies}

\clearpage

\bibliographystyle{IEEEtran}
\makeatletter
\renewcommand{\refname}{\vspace{-1.5ex}References}
\renewcommand{\IEEEbibitemsep}{-0.03\baselineskip}
\makeatother
\bibliography{ref}

\end{document}

%% file: sections/01_Introduction.tex
\section{Introduction}
\begin{figure} [!t]
    \centering
    \includegraphics[width=0.75\linewidth]{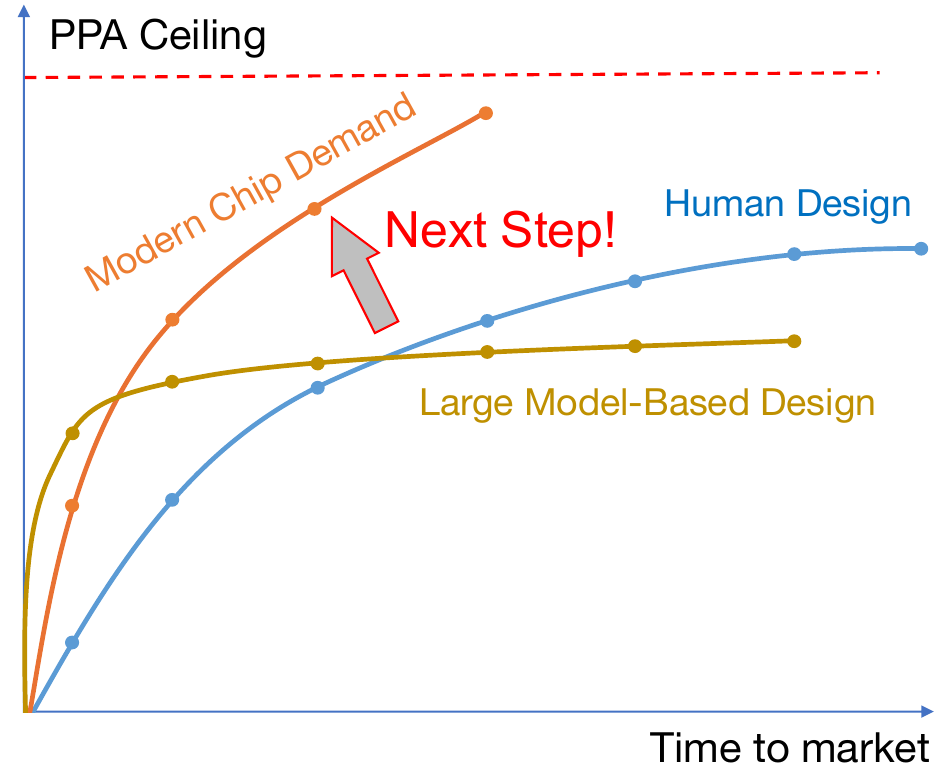}
    \vspace{-6pt}
    \caption{Qualitative Illustration of PPA Ceiling vs. Time-to-Market}
    \label{fig1}
    \vspace{-12pt}
\end{figure}

The continued scaling of integrated circuits (ICs) has led to designs of unprecedented complexity, with billions of transistors integrated into a single system-on-chip (SoC). This growth underpins a semiconductor market valued at approximately \$628 billion in 2024 and projected to exceed \$1 trillion by 2030~\cite{SIA2025,McKinsey2022}. At the heart of this growth, Electronic Design Automation (EDA), a \$15 billion industry~\cite{TBRC2025}, provides the software infrastructure and methodologies that make such designs feasible. For decades, advances in EDA algorithms and tool flows have kept pace with increasing design demands, enabling steady improvements in power, performance, and area (PPA). Today, those demands are further driven by application domains such as artificial intelligence, high-performance computing, and autonomous systems.
At the same time, conventional EDA approaches are running into fundamental limits. PPA targets tighten, design spaces grow exponentially, and human-crafted heuristics in existing tools can struggle with optimization. Fig.~\ref{fig1} illustrates the so-called ``PPA ceiling'', where iterative refinements tend to converge to local optima, leaving potentially better solutions unexplored. This can result in extended development timelines, increased engineering cost, and missed performance opportunities.

The search for better methods to navigate the design space has led to growing interest in artificial intelligence. By learning complex, non-linear patterns from vast datasets of prior designs, these models offer a fundamentally different approach to problem-solving—one capable of exploring the design space more holistically. Early work in ``AI for EDA'' applied classical and deep learning techniques to specific tasks such as placement, routing, or logic synthesis~\cite{huang2021machine}. Despite delivering measurable gains, these approaches often target point-solutions integrated into existing flows, with limited impact on the overall design cycle.

\begin{figure*} [!t]
    \centering
    \includegraphics[width=0.8\linewidth]{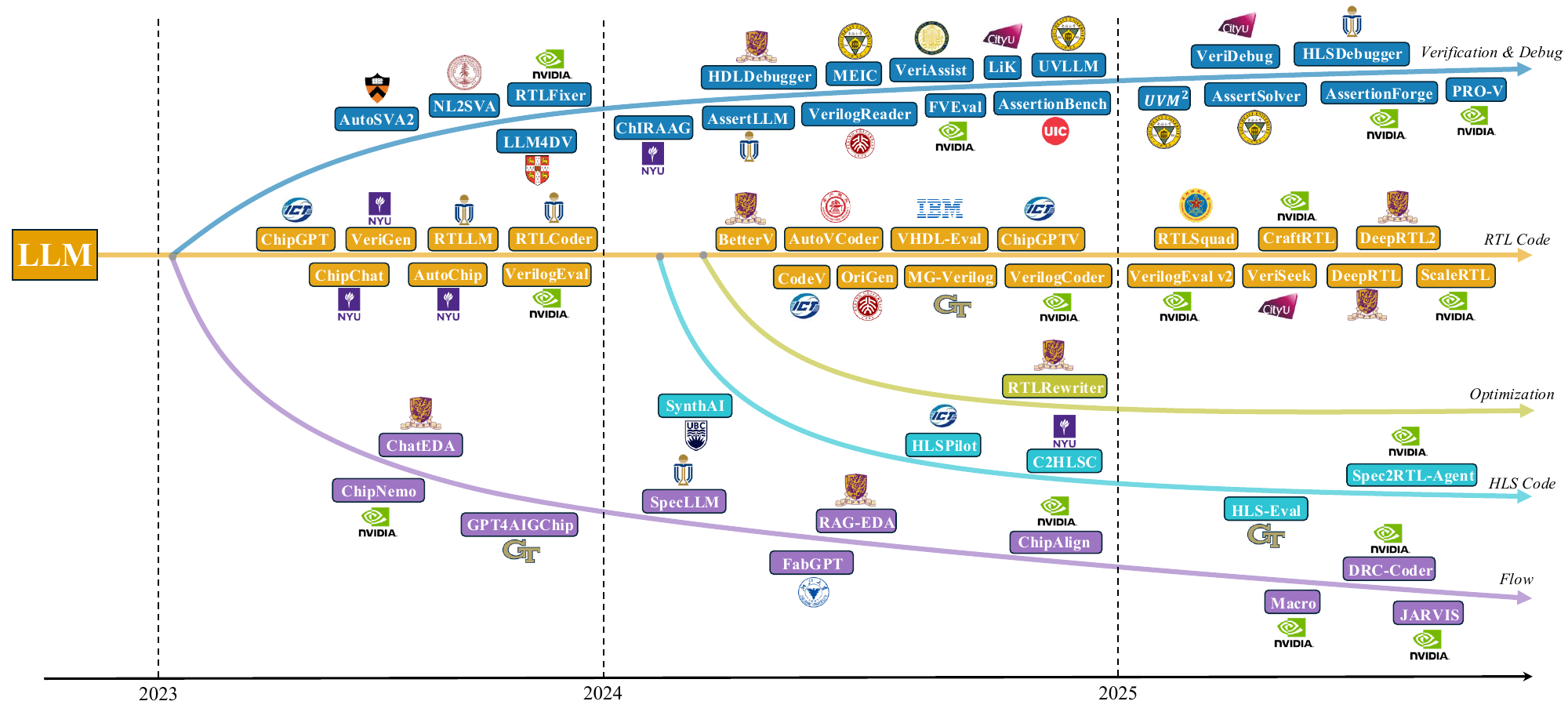}
    \vspace{-6pt}
    \caption{Large Language Models for Hardware Design}
    \label{fig:llm_overview}
    \vspace{-12pt}
\end{figure*}

The advent of large-scale and general models has offered a new path forward. 
Large Language Models (LLMs), trained on hardware description languages (HDLs) and design specifications, can interpret high-level intent and generate code~\cite{pan2025survey}. Applications include architectural design space exploration~\cite{wang2024chatcpu}, automated generation of verification components~\cite{xiao2024llm,hassan2024llm}, and direct synthesis of register-transfer level (RTL) descriptions from natural language~\cite{liu2024rtlcoder,ho2025verilogcoder}. In parallel, Large Circuit Models (LCMs) aim to learn from multi-modal circuit data—combining structural and functional representations from Verilog, netlists, and layouts—to support tasks such as functional understanding and physical optimization~\cite{chen2024large}.

Despite this progress, distinct technical challenges remain for both LLM- and LCM-based approaches. For LLMs, issues include maintaining functional correctness in generated code, accurately modeling hardware-specific constraints such as timing closure and design-for-test requirements, and achieving PPA outcomes that approach those of expert engineers. Their integration into downstream verification, validation, and sign-off stages is still limited, and the scarcity of high-quality training data for advanced technology nodes constrains their applicability in leading-edge processes. For LCMs, the main difficulties lie in representing and reasoning over heterogeneous circuit modalities—ranging from RTL and gate-level netlists to physical layouts—while preserving electrical, timing, and geometric fidelity. A key open problem is enabling these models to capture hierarchical design context and multi-modal interactions in a way that remains both computationally tractable and directly useful for downstream EDA tasks.

These challenges motivate a key challenge: designing and integrating large models into EDA workflows to reliably produce functionally-correct designs with competitive PPA, while maintaining robustness and trustworthiness needed for semiconductor manufacturing. Two model paradigms are relevant here: (1) LLMs interpret high-level human intent and automate complex coding and verification tasks, (2) LCMs learn from multi-modal circuit data to gain a structural and functional understanding of hardware, an ability essential for advanced PPA optimization.
Scaling existing models is not sufficient; we need advances in model architectures, domain-specific data representations, and tight coupling with rigorous verification flows. Research is therefore refocusing from \emph{whether} AI can assist in hardware design to \emph{how} it can be engineered into a dependable tool that consistently achieves high PPA without undermining production reliability.

\begin{figure*}
    \centering
    \includegraphics[width=0.8\linewidth]{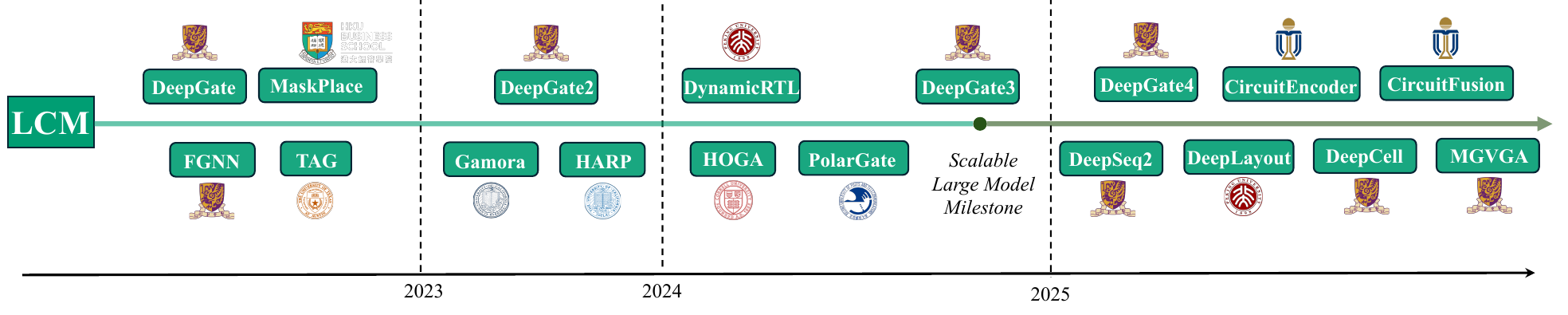}
    \vspace{-6pt}
    \caption{Large Circuit Models for Hardware Design}
    \label{fig:lcm}
    \vspace{-12pt}
\end{figure*}

This paper examines the current state and future prospects of large models for hardware design, synthesizing perspectives from the ICCAD 2025 panel discussion. Section~\ref{sec:background} provides background on current applications of LLMs and LCMs in hardware design. Section~\ref{sec:opportunities} outlines emerging opportunities, Section~\ref{sec:challenges} discusses integration challenges, and Section~\ref{sec:panel} summarizes key insights from the panelists. Section~\ref{sec:recommendations} presents recommendations for future research, and Section~\ref{sec:conclusion} offers concluding remarks.

\section{Background}
\label{sec:background}
\subsection{Large Language Models for Hardware Design}
LLMs are naturally relevant at design entry
to translate natural-language specifications into register-transfer level (RTL) code,  Verilog or VHDL.
Now their use is moving 
toward agentic, tool-in-the-loop pipelines that include verification. ChipGPT~\cite{chang2023chipgpt} establishes end-to-end translation from natural language to Verilog. Recent work  for RTL includes RTLCoder~\cite{liu2024rtlcoder} and introduces multi-agent planners that invoke simulators and waveform tracing, such as VerilogCoder~\cite{ho2025verilogcoder}, to improve correctness through feedback loops. Benchmarking is maturing: VerilogEval has advanced to v2
~\cite{pinckney2025revisiting} but is saturating (many LLMs achieve high scores), which is why it is being replaced by CVDP \cite{pinckney2025cvdp}.
Methodologically, BetterV~\cite{pei2024betterv} explicitly steers generation and improves controllability. Beyond generation, DeepRTL~\cite{liu2025deeprtl} formalizes RTL understanding tasks, and DeepRTL2~\cite{liu2025deeprtl2} proposes a versatile model that unifies generation- and embedding-based objectives at the RTL stage. In parallel, the HLS path is adopting multi-agent scaffolding to bridge the progression from natural language to C/C++ and then to HLS, as exemplified by SynthAI~\cite{sheikholeslam2024synthai}.

Beyond code generation, LLMs support verification and debugging. Assertion generation now targets full natural-language specifications, as in AssertLLM~\cite{yan2025assertllm}, while shared evaluations such as AssertionBench~\cite{pulavarthi2025assertionbench} provide consistent baselines and metrics. Practical repair pipelines, exemplified by RTLFixer~\cite{tsai2024rtlfixer}, address common failure modes in generated RTL and complement human-in-the-loop debugging.

Across the design flow, LLMs are increasingly adapted to the EDA domain and equipped with retrieval capabilities to improve faithfulness and practical utility. ChipNeMo~\cite{liu2023chipnemo} demonstrates how domain-adapted language models can better align with hardware design terminology and workflows, while RAG-EDA~\cite{pu2024customized} integrates targeted retrieval over tool documentation to reduce hallucination and improve traceability. Specialized agents automate rule-coding tasks, such as design-rule checks in DRC-Coder~\cite{chang2025drc}. Beyond drafting, LLMs also optimize RTL via search-and-rewrite strategies, as in RTLRewriter~\cite{yao2024rtlrewriter}. Fig.~\ref{fig:llm_overview} provides a detailed view of the progression of LLM applications in the hardware design process.

\subsection{Large Circuit Models for Hardware Design}

LLMs use natural-language processing in IC design, whereas Large Circuit Models (LCMs) strive for AI-native foundation models for EDA. LCMs must be  inherently multi-modal, encoding circuit semantics and structure across specifications, RTL code, netlists, and layouts to learn interactions between logical function, physical structure, and PPA. Unlike LLMs that linearize designs into text, LCMs must use  natively handle circuit topology, geometry, function, timing, etc~\cite{chen2024large}.

 The historical trajectory of LCMs (Fig.~\ref{fig:lcm}) begins with applying Graph Neural Networks (GNNs) for netlists, such as DeepGate~\cite{li2022deepgate}, which establishes core circuit representation learning. 
 In parallel, modality-specific models attempt to improve key stages of the design flow, such as placement with MaskPlace~\cite{lai2022maskplace}. A recent milestone, DeepGate3~\cite{shi2024deepgate3}, demonstrates scaling laws for circuit learning and provides a scalable training framework, conducive of LCMs and stronger downstream EDA performance.
The latest phase targets unified, multi-task models that blend complementary representations across design stages. DeepCell~\cite{shi2025deepcell} integrates pre- and post-mapping graphs to improve mapping engines and ECO, while CircuitFusion~\cite{fang2025circuitfusion} combines heterogeneous modalities for broader design reasoning. The shift from task-specific GNNs to large, multimodal foundation models marks the current frontier of LCM research.

%% file: sections/02_Opportunities.tex
\section{Opportunities}
\label{sec:opportunities}

Large models hold the potential to redefine design methodologies, enhance productivity, and accelerate innovation. Their integration into the EDA workflow presents major opportunities to address long-standing challenges in hardware design: design generation, faster verification, intelligent design- space exploration, and the democratization of EDA tools.

\subsection{High-Level Design Generation and Synthesis}
Automated RTL code generation from abstract high-level descriptions remains an attractive (but difficult-to-accomplish) application of LLMs because it can reduce manual coding effort. If sufficiently robust, it would allow engineers to specify design intent in English or Chinese, with subsequent translation into synthesizable Verilog or VHDL. IC designers can then focus on architectural innovation. To illustrate this approach, AutoChip~\cite{thakur2023autochip} refines LLM-generated code using tool-feedback-driven loops. This automation  could reduce time-to-market and lower the barrier to entry for IC design.

\subsection{Accelerating Hardware Verification}
Verification is widely recognized as the most time-consuming and resource-intensive phase of the chip-design cycle. Large models promise to alleviate this bottleneck. LLMs automate the generation of verification collaterals, such as testbenches and SystemVerilog Assertions (SVA), essential for both simulation-based and formal verification~\cite{qiu2024autobench, hu2024uvllm, qayyum2024ats}. By generating comprehensive test scenarios and properties from design specs, LLMs help teams attain coverage closure faster. This automation frees verification engineers from writing boilerplate code, so that they focus on advanced and corner-case verification strategies, thus improving the overall quality and correctness of the final design.

\subsection{Intelligent Design Space Exploration \& PPA Optimization}
The combinatorial complexity of the PPA design space strains conventional EDA tools. Here LCMs may offer unique advantages if, training on multi-modal data (logic, topology, and physical layout) reveals patterns of PPA optimization gives them a wholistic
understanding the intricate interplay between a circuit's functionality, topology, and physical geometry. To leverage such knowledge, they will additionally need to act as reasoning engines. Such LCMs could optimize micro-architectures and provide guidance for physical implementation. Unlike localized, heuristic-based optimization, we foresee holistic, data-driven design-space exploration. 

\subsection{Augmenting Core EDA Algorithms and Tool-chains}
Large models can serve as high-level interfaces, but they also promise to augment and enhance core EDA algorithms by embedding AI into the computational engines or control those in the tool-chain. An LLM could orchestrate complex sequences of operations within a synthesis or place-and-route flow. By analyzing logs from previous design runs, it could improve commands and parameters to achieve better PPA for a specific design, moving beyond static scripts to dynamic, context-aware optimization strategies (but this niche is well-explored by LLM-free optimization tools from Cadence and Synopsys, such as DSO.ai). LCMs may be more suited for such tasks if their native understanding of circuit topology and physical properties makes their early PPA predictions accurate enough to guide traditional EDA algorithms. For example, an LCM-based  cost model could
evaluate a placement without requiring time-consuming timing analysis.\footnote{Such functionality can be illustrated by Synopsys ChipArchitect, where it relies on more traditional ML techniques.}
Such deep integration can help AI-based EDA tools outweigh current SOTA.

%% file: sections/03_Challenges.tex
\section{Challenges}
\label{sec:challenges}

Despite their promise, integrating large models into production EDA workflows is fraught with formidable challenges. These hurdles span the entire lifecycle of model development and deployment, from data acquisition and representation to the fundamental requirements of reliability and trust. Addressing these issues is critical for moving beyond promising research demonstrations to creating tools that are robust and dependable enough for mission-critical silicon design.

\subsection{Reliability and the Spectre of Hallucination}
The adoption of large models in EDA has so far been hampered by high error rates, poor reliability, and non-determinism. LLMs are probabilistic models prone to {\em hallucinations} (confabulations) --- generating outputs that are syntactically plausible but functionally incorrect or nonsensical. In the context of IC design, correctness is paramount, so even a subtle error in generated RTL code or assertion can lead to catastrophic silicon failure, costly re-spins, and product delays. 
Thus, creative generative capabilities of LLMs must be tightly coupled with formal verification to check and mathematically prove correctness~\cite{qayyum2024ats,hassan2025coins}.

\subsection{The Semantic Gap: Representing Complex Circuit Data}

Natural languages (and thus LLMs) struggle to capture rich multi-modal circuit descriptions that blend
functionality (logic), topology (structure),  geometry (physical layout), timing, power, and other aspects. An even greater challenge is making this combined information available for large-scale high-performance optimization no worse than in existing EDA tools. A minor structural change in a netlist can have profound and non-local impacts on timing, power, and signal integrity. This causality is not natively captured by models trained on sequential text. Such mismatches motivate the development of multi-modal LCMs and AI-friendly circuit representations that facilitate effective capture, manipulation, and generation of gate-level IC designs. In this context, multi-modal reasoning and optimization pose additional frontiers.

\subsection{Data Scarcity, Privacy, and Security}
The performance of any large model is fundamentally dependent on the quality and quantity of its training data. In the EDA domain, this presents multiple challenges. {\bf First}, the amount of public IC design data is many orders of magnitude smaller than what is used to train natural-language models. {\bf Second}, the most complex SoCs designed for leading process nodes are highly proprietary and closely guarded. Even if EDA vendors have access to IC designs from multiple customers, work must be compartmentalized to prevent data leakage.
This scarcity of high-quality data in any one place prevents effective training of models that can generalize to new IC designs. {\bf Third}, transmitting sensitive data to third-party cloud-based EDA providers raises significant risks for intellectual property (IP). The industry is currently converging on acceptable solutions such as on-premise, locally deployed models and advanced security for cloud-based EDA. The paucity of training data remains a major showstopper.

\subsection{Explainability and Building Trust}
Models that operate as inscrutable "black boxes" obstruct adoption of AI-driven tools and trust in them. Validating a suggested architectural change or a bug fix is much easier when underlying reasoning is clear or at least fully spelled out. Thus, explainability is a major challenge for many AI models with complex and opaque decision-making, .and especially for models that lack an explicit reasoning process. Future research seeks models with traceable, logical explanations for their outputs. Otherwise, AI may be limited to advisory assistance and not become a trusted design partner.

%% file: sections/04_ExpertOpinions.tex
\section{Expert Opinions and Insights}
\label{sec:panel}
This panel brings together leading experts who, while enthusiastic about AI's transformative potential in EDA, offer distinct perspectives on the path forward. Their discussion tackles immediate challenges like LLM hallucinations and data privacy, but centers on a key question of model suitability: are general-purpose LLMs sufficient, or does circuit data's unique complexity demand specialized LCMs? Despite differing views, a shared imperative emerges for reliability and correctness, underscoring the need to integrate any AI-driven innovation with rigorous verification. The following sections organize these insights by key themes.

\subsection{LLM vs. LCM: Generalization or Specialization?}

\textbf{Grace Li Zhang} emphasizes that LLMs are not universally applicable to all tasks in the EDA workflow. 
Instead, different models may be better suited for specific tasks, and understanding their strengths and limitations is essential to unlocking the full potential of AI in hardware design.

\textbf{Xi Wang} proposes a holistic \emph{LLM-Aided Design} framework, which reimagines hardware development beyond simply enhancing existing EDA tools. It envisions LLMs as central to a unified, adaptive design methodology that scales with increasing complexity.
A key advantage of this approach is its potential to automate the entire design stack, especially in stages traditionally requiring deep domain expertise—such as performance modeling, RTL implementation, and functional verification. LLMs can significantly reduce the time and resources required for tasks like the construction of testbench, SystemVerilog Assertions (SVA), and reference models. Projects such as ChatCPU~\cite{wang2024chatcpu}, ChatModel~\cite{ye2025chatmodel}, MEIC~\cite{xu2024meic}, UVLLM~\cite{hu2024uvllm}, and VeriDebug~\cite{wang2025veridebug} have demonstrated this capability, showing how LLMs can automate complex tasks, streamline verification, and accelerate the overall design process.
However, Wang notes that realizing this vision requires a breakthrough in circuit representation, as current formats are poorly aligned with the processing mechanisms of LLMs. Developing AI-friendly representations is thus critical to the success of LLM-driven methodologies.

\textbf{Qiang Xu} 
underscores the fundamental differences between circuit data and natural language. Circuit representations inherently combine functionality (logic), topology (structure), and geometry (layout), making them highly sensitive to minor modifications. A minor change in a netlist's structure can trigger serious non-local effects in circuit function --- a complexity not captured by general-purpose LLMs trained on text. This is why we need LCMs that can reason and learn to optimize IC designs for power, performance, and area (PPA), while ensuring design correctness.
Rather than viewing LCMs as a replacement for LLMs, collaboration is seen as more promising. Xu suggests a division of labor that plays to the strengths of each model—the synergy between LLMs and LCMs: the ``What'' vs. the ``How''
\begin{itemize}
    \item \textbf{LLMs for the ``What'':} LLMs excel at interpreting human intent. They are ideally suited for the ``front door" of the design process—translating a high-level, unstructured requirement (\emph{e.g.}, designing a low-power RISC-V core for IoT devices) into a formal specification or a high-level architectural description. The LLM acts as the ultimate natural language interface.
    \item \textbf{LCMs for the ``How'':} Once the design intent is captured, the LCM takes over. It acts as the ``expert engineer", translating the specification into an optimized and correct gate-level netlist with its deep understanding of structure. The LCM would navigate the complex PPA trade-offs, suggest micro-architecture changes, and provide guidance for physical implementation --- tasks that require a native understanding of circuit principles.
\end{itemize}

\begin{figure} [!t]
    \centering
    \includegraphics[width=0.88\linewidth]{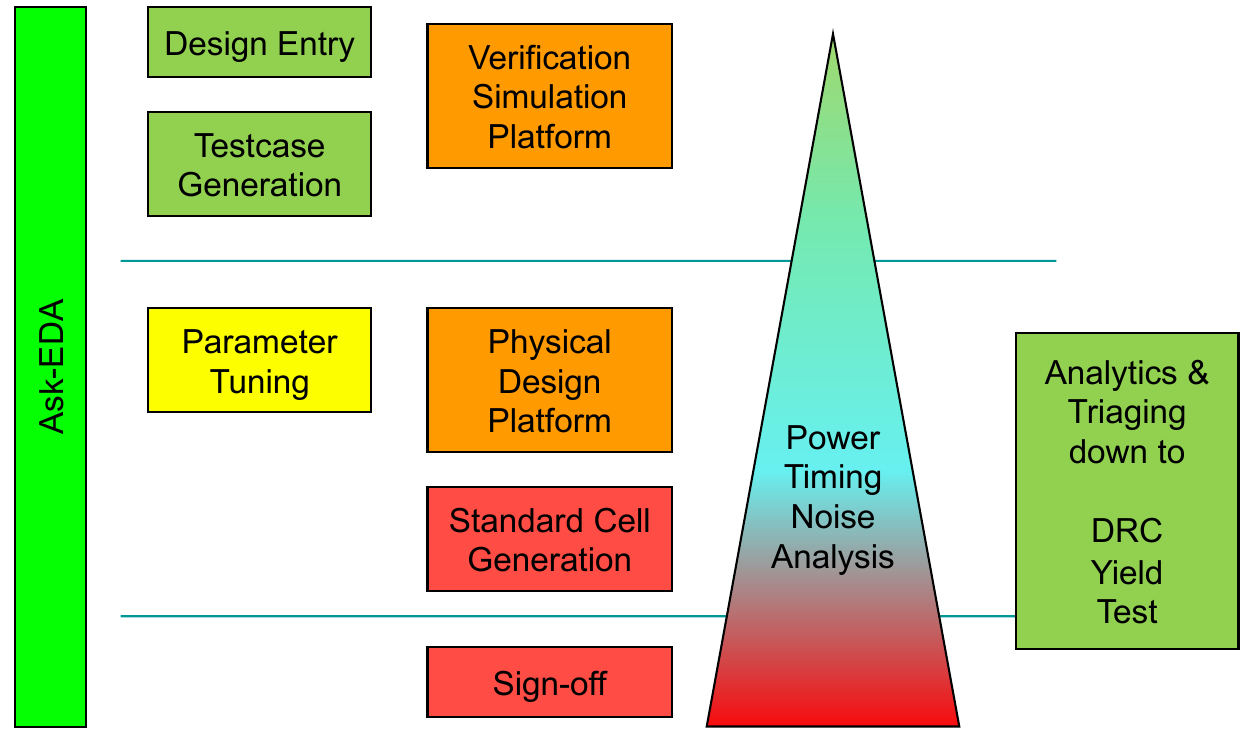}
    \vspace{-6pt}
    \caption{Applicability of generative models across the EDA workflow}
    \label{fig:leon_stok}
    \vspace{-10pt}
\end{figure}

\textbf{Leon Stok} reinforces this division of labor by pointing to practical use cases where LLMs are already proving valuable~\cite{zhao2025chain}. 
Today’s EDA tools are deeply complex, with workflows requiring extensive configuration across various PDKs and IPs. LLM-powered assistants, such as askEDA~\cite{shi2025improving}, help users navigate these flows by leveraging RAG or RAFT methods to answer domain-specific queries. When combined with hardware-trained LLMs~\cite{dupuis2025customizing}, this enables intuitive interfaces for configuring flows, understanding documentation, and generating test cases, especially when enriched with historical bug data.
Stok also notes that LLMs are particularly effective at managing the overwhelming output of EDA tools—such as timing reports, power analysis, or DRC violations --- by helping designers extract insights and identify critical bottlenecks. 
However, Stok cautions that the lack of precision in AI models currently limits their use in final sign-off stages, where 100\% accuracy is essential for billion-transistor designs.
Nonetheless, he sees significant promise in integrating AI earlier in the flow. Approximate models and reinforcement learning (RL) techniques can guide design-space exploration and parameter tuning, offering favorable trade-offs between accuracy and runtime. By combining RL with agentic LLM frameworks, more granular and intelligent optimization strategies may be unlocked, bridging the gap between exploration and execution. 

Stok concluded his ICCAD 2024 keynote, titled \emph{From Idea to IC: The Future of Chip Design}, with a heatmap illustrating where GenAI and agentic systems are most likely to impact EDA, as shown in Fig.~\ref{fig:leon_stok}. The greener the box, the greater the potential for near-term AI integration—a visual that underscores the opportunities for applying AI-driven methods well before final sign-off.

\textbf{Igor L. Markov} highlights both the promise and limitations of foundational models like LLMs in hardware design, advocating for a more nuanced perspective grounded in data types, model capabilities, and domain-specific constraints.
He emphasizes that the effectiveness of foundational models depends heavily on the nature of the data used for their training. For example, the recent success of TabPFN~\cite{hollmann2025accurate} demonstrates how synthetic causal numerical data --- generated from random shallow neural networks --- can enable breakthrough performance in tabular classification tasks. This strategy offers several benefits over relying on real-world data: it avoids copyright concerns, enables better quality control, and provides abundant, label-rich training samples.

Markov points to serious limitations of LLMs in numerically-intensive tasks. Standard transformer models are generally inefficient at representing and reasoning over high-precision numerical values. They struggle with even basic arithmetic operations~\cite{mcleish2024transformers} and fail to meet the accuracy demands of core EDA workflows like physical synthesis, placement, and timing analysis. Although tool‑augmented LLMs can invoke external tools (e.g., calculators or Python interpreters), such integration incurs performance penalties. LLMs find it challenging to determine when invoking an external tool is beneficial. To this end, LCMs should adopt architectures better suited to numerical data.

Reasoning-focused LLMs explore chains- or trees-of-thought. These models improve transparency and attempt to execute algorithms step by step, supporting some aspects of deductive symbolic reasoning. However, Markov cautions that the efficiency of {\em learned algorithms} remains far behind {\em domain-optimized implementations}. Recent findings reveal sharp
declines in output quality as task size increases, calling into question the analogy between LLM reasoning and human thinking~\cite{shojaee2025illusion}. 
Markov sees no reason to expect LCMs to be any better than LLMs in this aspect, whereas chain-of-thought reasoning is not yet fully-developed for LCMs.

Ultimately, Markov argues that key limitations create a protective moat around traditional algorithmic approaches in formal verification, logic synthesis, and physical design. While LLMs may play supportive roles --- particularly in early-stage specification tasks,
test generation, and assertion synthesis,  --- they fall short in the high-precision, large-scale computations that dominate backend flows.
He also tempers the hype around code-generation beyond coding assistants. While LLMs are increasingly used for software development, studies reveal a mismatch between perceived and actual productivity gains~\cite{metrref}, with many engineers rewriting or discarding generated code because, even when functionally correct, it is difficult to work with. Code assistants face even greater hurdles in IC design, where stringent constraints apply, multiple modalities are tightly coupled, and training data is scarce.

On the positive side, Markov acknowledges that LLMs are already in commercial use for targeted tasks: converting natural-language specs into structured formats, generating localized RTL code snippets and EDA scripts, and assisting with documentation and verification. Agentic LLMs capable of invoking specialized tools represent a promising direction for enabling practical, hybrid workflows in EDA.
\subsection{Reliability, Hallucinations, and Verification}

\textbf{Grace Li Zhang} identified hallucinations in LLM outputs as a critical barrier to industrial adoption in EDA. She emphasized that hallucinations are not merely factual inaccuracies but manifestations of a deeper issue: the lack of transparent reasoning in LLM decision-making. Understanding how LLMs arrive at specific outputs—step by step—is still an open research challenge. Zhang called for advancing interpretability techniques that shed light on LLM internal logic and for secure deployment strategies, such as on-premises inference or encrypted communication for cloud-based models, to protect sensitive design IP. Despite these risks, she maintained that LLMs hold transformative potential for streamlining design workflows—if reliability can be ensured.

\textbf{Rolf Drechsler} focused on the challenge of correctness in safety-critical silicon design. He noted that while LLMs have shown promise in automating verification-related tasks such as testbench generation~\cite{qiu2024autobench, qiu2025correctbench}, they remain inherently probabilistic and prone to hallucinations. In contexts where even a 0.1\% error rate could result in catastrophic failures, Drechsler argued that formal verification remains indispensable. He proposed a hybrid approach that tightly integrates LLMs with formal methods, allowing for creative synthesis paired with deterministic validation. Explainability, he argued, is the key to bridging the creativity of LLMs with the strict correctness guarantees required in EDA, and he called for robust, multi-dimensional evaluation metrics that capture both reliability and generative usefulness.

\textbf{Leon Stok} echoed these concerns about correctness, especially during sign-off. While he acknowledged the utility of LLMs in pre-sign-off phases—such as analyzing large volumes of timing, power, or DRC reports—he emphasized that final designs demand 100\% accuracy across billions of transistors and wires. In such stages, the tolerance for LLM error is effectively zero. Stok noted that LLMs are valuable for navigating complexity earlier in the flow, particularly when combined with agentic frameworks and historic bug data to assist in debugging and test generation. However, he drew a firm line at simulation performance and sign-off, where deeply optimized, deterministic algorithms remain irreplaceable.

\textbf{Igor L. Markov} notes that hallucinations increase with insufficient training data, limited access to verifiers, and inadequate reasoning abilities. Broad-interest natural-language data, especially in English, is abundant. But adequate HDL data and even EDA documentation are limited. Studies show that LLMs trained on limited amounts of data (including synthetic data/code) produce too many hallucinations to be useful, particularly when nuanced or domain-specific reasoning is required. In many cases, the Retrieval-Augmented Generation (RAG) framework helps LLMs ground their outputs in high-quality external sources, supports dynamic document updates without retraining, and allows LLMs to focus on most relevant documents, even include references to source data in responses. Particularly important is the continuous cycle of evaluation and document refinement supported by RAG, which addresses frequently recurring queries and their variants.
When combined with verification, RAG enhances factual accuracy, reduces hallucinations, and enables modular, domain-specific deployments. These enhancements apply to both LLMs and LCMs in principle. While LCMs remain further out, LLMs have become valuable for EDA tasks that rely on information retrieval. Strengthened with knowledge graphs, such LLM-based products are already in broad use by customers of EDA tools, as illustrated by ($i$) Synopsys {\sc Knowledge Assistant} which answers technical questions based on documentation with text and graphics, as well as ($ii$) Synopsys {\sc Run Assistant} that produces small Tcl and Python scripts by request, based on legacy scripts. Information retrieval for software can be illustrated by the {\sc GitHub Copilot} used for code completion used by numerous GitHub users.

{\em Reasoning} LLMs can follow simple algorithms, such as solving basic mathematical problems, and generate unit tests for software. Inference-time compute helps LLMs perform low-depth search to find better answers. Such advanced code LLMs from Anthropic and OpenAI are increasingly popular for {\em vibe coding}, especially for simple computer games and web site design. Derived applications {\sc Cursor}, {\sc Replit}, and {\sc Lovable} create hundreds of thousands applications per day from specs by people who have not coded before. This development process bottlenecks in diagnosing and fixing bugs.

Markov also noted that LLMs can be extended to invoke external tools: calculators, formal solvers, and full-fledged EDA tools. Advanced LLMs for IC design find use in the Synopsys {\sc Euclide IDE} developed for chip designers and verification engineers dealing with SystemVerilog and UVM. {\sc Euclide} accelerates their work by providing on-the-fly incremental compilation, as well as real-time bug detection, code quality improvement, and context-specific autocompletion, while linking with established verification tools. 

These expert insights collectively paint a multifaceted picture of how large models are reshaping the future of EDA, where optimism is tempered by critical concerns over data and correctness. The central debate is therefore not if AI will be used, but how the generative potential of LLMs can be integrated with domain expertise and what role specialized LCMs will ultimately play in this emerging ecosystem.

\vspace{-2pt}

%% file: sections/05_Future.tex
\section{The Way Forward}
\label{sec:recommendations}
\vspace{-3pt}
The integration of LLMs and LCMs into EDA offers a significant opportunity to accelerate chip design, but substantial challenges in reliability, efficiency, and scalability must be addressed. To realize this potential, future research must overcome these obstacles. 
Panelists shared diverse perspectives on the key directions that will shape the future of LLM-aided design, including improving model reliability, developing tailored circuit representations, advancing hybrid verification systems, and fostering deeper collaboration between academia and industry. The following recommendations offer a roadmap for advancing LLM and LCM integration into EDA, ensuring that these technologies reach their full potential in reshaping the hardware design landscape.

\textbf{Grace Li Zhang} emphasized that future research should focus on improving the reliability and explainability of LLMs to make them more dependable for hardware design tasks. Understanding and mitigating hallucinations is crucial, and more work is needed to explore how LLMs reason through decisions and how to ensure traceable logic in their outputs.

Regarding privacy and security, Grace recommended that local server deployments with encrypted communication to the cloud be explored further to balance data security with the computational flexibility needed for complex tasks.

Grace also suggested a more targeted approach in applying large models to specific areas of the EDA workflow. Rather than expecting a one-size-fits-all solution, it is important to identify the best-suited tasks for each type of large models, ensuring that their application is both effective and efficient.

\textbf{Xi Wang} offered several recommendations for advancing LLM-Aided Design:
\begin{itemize}
    \item \textbf{Collaborative Development of Benchmarks:}
    Current benchmarks are too simplistic for real-world hardware. To properly evaluate and improve AI models, academic and industry partners must collaborate on comprehensive, open-source benchmarks (e.g., GenBen~\cite{wan2025genben} and FIXME~\cite{wan2025fixme}) that can assess complex tasks like timing analysis, PPA, and advanced verification.
    \item \textbf{Advancing Human–GenAI-Friendly Design Representations:}
    To enrich semantic expressiveness while maintaining human interpretability, new design representations should combine traditional HDLs with graph-based IRs. This structure, when paired with interactive guidance and real-time feedback, will enable generative models to iteratively refine designs and produce higher-quality outputs.
    \item \textbf{Shifting from ``LLM for EDA'' to ``EDA for LLM'':}
    To realize AI's full potential, the community must shift from merely enhancing existing tools with AI to building new EDA tools explicitly for AI. This requires co-designing EDA algorithms with AI architectures to create AI-native systems capable of autonomous design, fostering an ecosystem where AI models drive, rather than merely assist, innovation.
\end{itemize}

\textbf{Qiang Xu} proposed two recommendations to meaningfully integrate both LLMs and LCMs into EDA. A primary recommendation is to accelerate research into foundation models that are native to the EDA domain. The community must move beyond adapting models from other fields and invest in building LCMs from the ground up. This requires a dedicated effort to create large-scale, high-quality, multimodal datasets that encompass the entire design lifecycle, from specifications to layouts (e.g.~\cite{li2025deepcircuitx, shi2025forgeeda}). Overcoming the challenge of proprietary data will necessitate community-driven platforms for data sharing and the development of sophisticated data augmentation and synthesis techniques. Furthermore, research should target novel neural architectures tailored for the unique hierarchical and graphical nature of circuits. The training objectives for these models must also evolve, shifting from narrow predictive tasks to broad, self-supervised goals that teach the model the fundamental principles of logical equivalence, timing closure, and PPA optimization, creating a true ``AI-rooted" foundation. 

The second major directive is to engineer the synergy between LLMs and LCMs to create a new class of intelligent EDA tools. In this model, LLMs serve as a conversational front-end to interpret a designer's intent from natural language, while LCMs act as the back-end reasoning engine to guide the complex implementation and optimization. This entire process, however, must be anchored by a commitment to trust: any generative output must be rigorously validated by traditional formal methods to ensure correctness.

\textbf{Rolf Drechsler} suggested that the future of LLMs in hardware design lies in the development of hybrid systems that combine the strengths of LLMs with formal verification tools. He recommended that LLMs be paired with formal reasoning engines to ensure that results are not only creative but also verifiable and reliable. This integration would enable traceable reasoning, allowing verification engineers to trust the outputs of AI-driven tools, even in complex or safety-critical scenarios.

Additionally, Drechsler called for the establishment of robust evaluation metrics to assess LLMs' capabilities in the context of EDA. These metrics would help the community distinguish between genuine advancements and hyped results, fostering a more scientific and repeatable approach to integrating LLMs into hardware design.
He emphasized that future research should focus on building reliable systems that the industry can trust. This would involve refining LLM architectures to reduce randomization issues and ensure that models produce accurate and verifiable results in critical design tasks.

\textbf{Leon Stok} emphasized that some of the most insightful guidance for integrating AI into engineering workflows comes from Ethan Mollick’s book \emph{Co-Intelligence: Living and Working with AI}. He noted that the principles outlined in the book are highly relevant to the EDA and chip design domains, offering a practical framework for safely and effectively incorporating AI tools and agents into existing processes.

The key principles include:

\begin{itemize}
    \item \textbf{Always include AI in the conversation:} Treat AI as a collaborator that can contribute ideas, insights, and support throughout the design process.
    \item \textbf{Maintain a human-in-the-loop approach:} Ensure human oversight remains central to decision-making, rather than relying solely on autonomous systems.
    \item \textbf{Assign the AI a persona and actively engage with it:} Give the AI a role or identity to foster clearer interaction, and challenge it to refine its outputs.
    \item \textbf{Assume this is the least capable AI you will ever use:} Set expectations accordingly, and stay prepared for rapid advancements and improvements in AI capabilities.
\end{itemize}

\textbf{Igor L. Markov} noted that EDA vendors are concerned about both the maturity and time-horizon of any new technology considered for implementation in commercial tools. To this end, he compared the adoption of ($i$) LLMs for natural language, software, and IC design, and ($ii$) future LCMs.

LLMs for code are now actively deployed and evaluated, even as the long-term impact of coding assistants on engineer productivity remains unclear~\cite{metrref}.
In contrast, LCMs are currently far from achieving comparable maturity and adoption levels. This lag is largely due to fundamental gaps in support of circuit-related modalities such as large graphs. Discriminative and other observational tasks, such as timing and power analysis, tend to benefit from GNNs and can perform learning separately {\em for each graph}, but demanding generative tasks for large-scale gate-level circuits require AI capabilities not available today. Key research directions include developing highly accurate, AI-friendly representations of large sparse graphs (likely vector embeddings), enabling learning of large-scale structure (beyond mere texture) across graphs of varying sizes, creating more scalable Graph Transformers or comparable techniques, and supporting graphs with diverse node attributes (including functional, geometric, timing, and power information). Filling in these gaps for the graph modality can unlock significant future developments on LCMs. Future techniques will need to support handling large amounts of numerical data efficiently, as well as integrated reasoning (not just tool invocation) and large-scale optimization.

Significant progress in AI for EDA is only possible on a solid base of empirical validation. The well-known pitfall of {\em data leakage} can produce misleadingly good results when AI is evaluated on data similar to training data. In practice, many benchmark sets {\em saturate} over time as various AI tools reach higher scores, sometimes by absorbing similar training data without improving real-world application performance. This issue can be addressed by describing training data well and through evaluations performed by independent third parties. 
\vspace{-3pt}

%% file: sections/06_Conclusion.tex
\section{Conclusion}
\label{sec:conclusion}
\vspace{-1pt}

Large models are at once a disruptive force and a significant unknown in hardware design. Their power to automate and innovate is undeniable, but reliability and trust concerns hinder adoption. The path from hype to revolution requires anchoring the creative capabilities of AI in the bedrock of formal verification. Whether these models become indispensable partners or remain specialized tools is the defining question for the next era of EDA—an answer that will be forged through both ambitious research and pragmatic engineering.

\vspace{-3pt}

%% file: sections/Biographies.tex
\section*{Panelist Biographies}
\textbf{Qiang Xu}
is a Professor at The Chinese University of Hong Kong. He previously served as the Chief Scientist of the National Technology Innovation Center for EDA, China. In 2024, he co-initiated the concept of Large Circuit Models (LCMs), advocating for AI-native EDA solutions. He has published over 200 papers, earning several Best Paper awards at top-tier conferences, an ICCAD Ten-Year Retrospective Most Influential Paper Award, and an AAAI Most Influential Paper.

\vspace{5pt}

\textbf{Leon Stok}
is Vice President of EDA at IBM. With over 30 years in Design Automation, his work has spanned high-level synthesis to placement-driven synthesis and prescriptive layout design. He has developed numerous EDA tools and has been a frequent keynote speaker and panelist at major conferences. Dr. Stok has served in many roles on the DAC executive committee, including chairing the 48th Design Automation Conference, and is an IEEE Fellow. 
\vspace{5pt}
 
\textbf{Rolf Drechsler}
is a Professor at the University of Bremen and Head of the Cyber-Physical Systems Group at the German Research Center for AI (DFKI). His research focuses on circuits and system design. He has served as General Chair for the IEEE European Test Symposium (ETS) and the IEEE/ACM Internationa Conference on Computer-Aided Design (ICCAD). He is a recipient of two 10-year Retrospective Most Influential Paper Awards at ASP-DAC and the Best Paper Award at ICCAD, respectively. He is an ACM and IEEE Fellow.
\vspace{5pt}

\textbf{Xi Wang}
is an Associate Professor at Southeast University, where his research explores the intersection of LLMs and EDA for agile hardware design. He introduced ChatCPU, an LLM-assisted platform for agile CPU design and verification, which was nominated for the Best Paper Award at DAC 2024. His subsequent work, ChatDV, automates functional verification and debugging. Dr. Wang’s research has been recognized with awards including the IEEE IPDPS 2021 Best Paper Award and the ISSCC 2023 Code-A-Chip Award. 
\vspace{5pt}

\textbf{Grace Li Zhang} is a Professor at TU Darmstadt leading the Hardware for Artificial Intelligence group. Her research encompasses efficient machine learning, hardware-software architectures, hardware accelerators for AI algorithms, and neuromorphic computing. She has received Best Paper Awards/Nominations at conferences such as DAC, DATE, MLCAD and ISQED. 
\vspace{5pt}

\textbf{Igor L. Markov}
is a Distinguished Architect at Synopsys, leading the AI Disruption Task Force.
He previously worked at Google on search, at Meta on AI, and as an EDA professor at the University of Michigan. His work focuses on algorithms, software, and AI methodologies for IC design, as well as quantum computers. He is an IEEE Fellow and an ACM Distinguished Scientist. Dr. Markov has co-authored five books and over 200 refereed publications, some of which received Best Paper awards at DATE, ISPD, ICCAD, and TCAD.